\documentclass[letterpaper]{article} 
\usepackage{aaai2027}  
\usepackage[hyphens]{url}  
\usepackage{graphicx} 
\usepackage{natbib}  
\usepackage{caption} 
\usepackage{algorithm}
\usepackage{algorithmic}

\usepackage[utf8]{inputenc} 
\usepackage[T1]{fontenc}    
\usepackage{url}            
\usepackage{booktabs}       
\usepackage{amsfonts}       
\usepackage{nicefrac}       
\usepackage{microtype}      
\usepackage{xcolor}         
\usepackage{amsmath}
\usepackage{enumerate}
\usepackage{amssymb}
\usepackage{pifont}
\usepackage{graphicx}
\usepackage{multirow} 

\usepackage{newfloat}
\usepackage{listings}
\DeclareCaptionStyle{ruled}{labelfont=normalfont,labelsep=colon,strut=off} 
\floatstyle{ruled}
\newfloat{listing}{tb}{lst}{}
\floatname{listing}{Listing}

\usepackage{booktabs}

\newcommand{\err}[1]{{\scriptsize(#1)}}
\newcommand{\best}[2]{{\bfseries #1{\scriptsize(#2)}}}

\title{HySTAR: Anchored Hypergraphs for Stable Credit Assignment in Cooperative Multi-Agent Reinforcement Learning}
\author {
    Xinglong Luo\textsuperscript{\rm 1}\equalcontrib, 
    Yuding Zhang\textsuperscript{\rm 1}\equalcontrib, 
    Yuheng Kuang\textsuperscript{\rm 1}, 
    Shuxuan Yuan\textsuperscript{\rm 1}, 
    Zhenni Zeng\textsuperscript{\rm 1}, \\
    Weiqiang Zhu\textsuperscript{\rm 2}, 
    Zhenhai Ji\textsuperscript{\rm 2}, 
    Zhengning Wang\textsuperscript{\rm 1}\corresponding
}
\affiliations {
    \textsuperscript{\rm 1}University of Electronic Science and Technology of China  
    \textsuperscript{\rm 2}Independent Researcher\\
}

\begin{document}

\maketitle

\begin{abstract}
Cooperative multi-agent reinforcement learning under partial observability and shared rewards requires assigning team outcomes to individual agents and high-order coalitions. A MAPPO-style critic compresses joint behavior into one global value, while critics that dynamically reconstruct the grouping topology change the mapping from agents and coalitions to value components as interactions or active agents evolve. We refer to this inconsistency as structural target drift. We introduce HySTAR, a MAPPO-based framework that separates adaptive representation learning from a temporally consistent high-order value-decomposition basis. HySTAR anchors an overlapping sparse hypergraph as a uniformly covered decomposition scaffold, uses a spatiotemporal encoder to represent physical and task-dependent interactions, and combines temporal and structural relevance to construct agent-specific advantages. Experiments on SMAC, GRF, Traffic Junction, and MPE demonstrate consistent improvements over MAPPO-style, value-factorization, and dynamic-grouping baselines. On the hardest SMAC settings, HySTAR achieves relative gains of 16.7\% over MAPPO and 15.6\% over HYGMA, ranks first on all six GRF scenarios, reduces Traffic Junction convergence epochs by up to 40.2\% relative to MAGIC, and obtains the highest MPE episode rewards. Controlled topology, agent-death, neighborhood, and parameter analyses support the benefit of anchoring the decomposition scaffold while adapting the propagated representations.
\end{abstract}

\section{Introduction}
\label{sec:intro}

\begin{figure*}[t]
    \centering
    \includegraphics[width=1\linewidth]{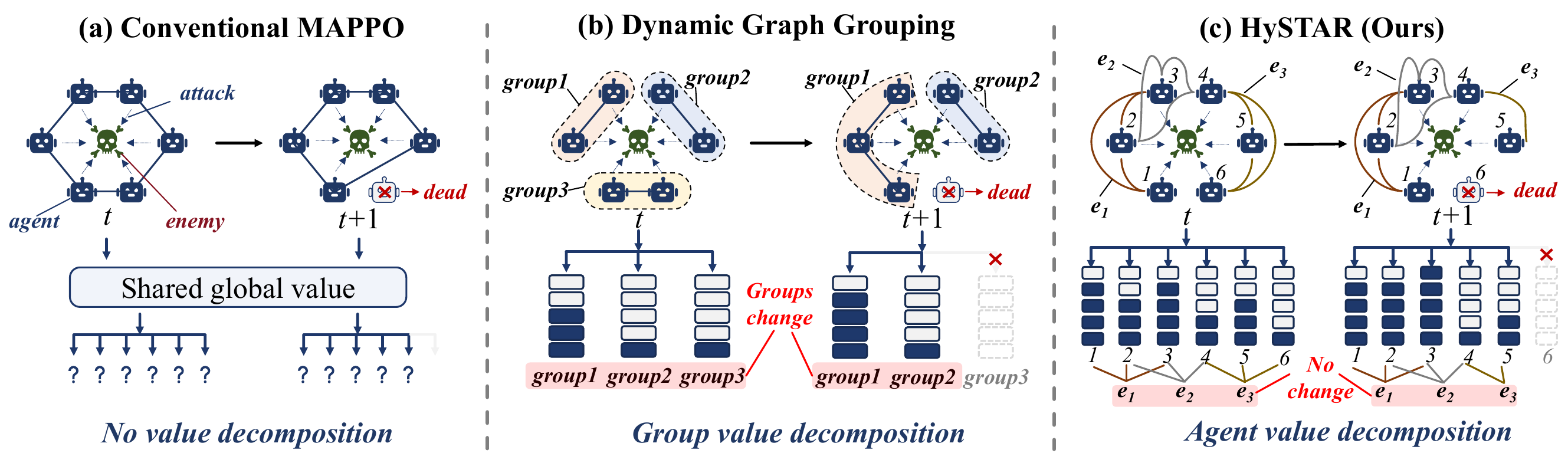}
    \caption{
        Motivation and design of HySTAR. MAPPO uses a single global value; dynamic grouping introduces group-level credit and changes its basis when relations or the active-agent set evolve; HySTAR maintains a stable agent-level decomposition basis through an anchored sparse hypergraph while adapting the representations propagated over it.
    }
    \label{fig:teaser}
\end{figure*}

Cooperative Multi-Agent Reinforcement Learning (MARL) commonly adopts Centralized Training with Decentralized Execution (CTDE) under partial observability and shared rewards~\cite{yu2022surprising,sunehag2017value}. Its central difficulty is credit assignment: a shared return indicates team success while leaving the contributions of individual agents and high-order coalitions unresolved. As illustrated in Fig.~\ref{fig:teaser}(a), MAPPO produces one global value and therefore represents coordinated contributions---such as focus fire, covering, and complementary role execution---only implicitly.

Relational critics introduce structured value components, while dynamic grouping introduces a second source of instability. Fig.~\ref{fig:teaser}(b) shows that agent death and relation changes merge, split, or remove groups, so both the representations and the decomposition basis of the critic change. We refer to this inconsistency as \emph{structural target drift}: the mapping from agents and coalitions to value components varies across timesteps even when the semantic coordination role remains similar. The resulting design requirement is a stable credit-assignment basis together with representations that model time-varying interactions.

We propose \textbf{HySTAR}, short for \textbf{Hy}pergraph-anchored \textbf{S}patio\textbf{T}emporal \textbf{A}gent \textbf{R}epresentation, to decouple these two requirements. As shown in Fig.~\ref{fig:teaser}(c), HySTAR anchors an overlapping sparse hypergraph as a persistent agent-level value-decomposition scaffold, while the ST-Encoder adapts node representations to current spatial interactions and temporal trajectories. A regular overlapping hypergraph provides a temporally consistent and uniformly covered index structure, while the ST-Encoder and learned messages encode physical and task-dependent interaction content. Its sparse incidence pattern supports multi-hop high-order aggregation with only $\mathcal{O}(N(k+1))$ entries. AHVD evaluates agent and team values over this scaffold, while STCA combines temporal and structural relevance to construct agent-specific advantages. This division of responsibility is central to HySTAR: the anchored structure stabilizes the support used for value decomposition, whereas the learned representations retain the flexibility needed to describe changing observations, roles, and interactions. HySTAR therefore follows the principle \emph{anchor the topology, adapt the representations} within a communication-augmented CTDE setting, with privileged global state used only during training.

The experiments examine this design from complementary perspectives. HySTAR obtains the best reported result on 11 of 12 selected SMAC maps. On \texttt{3s5z\_vs\_3s6z}, it improves over MAPPO from $84.4\%$ to $98.5\%$, a 16.7\% relative gain; on \texttt{6h\_vs\_8z}, it improves over dynamic-grouping HYGMA from $85.0\%$ to $98.3\%$, a 15.6\% relative gain. Traffic Junction further evaluates changing active-agent sets, where HySTAR reaches $99.9\%$ success and converges earlier than HYGMA and MAGIC. The highest final rewards on all three MPE tasks show that anchoring also benefits fixed-team coordination. HySTAR additionally ranks first on all six GRF scenarios, with relative gains reaching 14.3\% over MAPPO and 33.4\% over HYGMA. Controlled topology and agent-death analyses further show lower stability for inactive-agent masking alone.

Our contributions are:
\begin{enumerate}[1)]
    \item a communication-augmented CTDE framework that separates a persistent credit-assignment basis from adaptive spatiotemporal representations;
    \item AHVD and STCA for anchored high-order value decomposition and agent-specific credit shaping under shared rewards; and
    \item extensive evaluations across four MARL benchmarks, together with topology, agent-death, neighborhood, and parameter-efficiency analyses.
\end{enumerate}

\section{Related Work}
\subsection{Multi-Agent Reinforcement Learning}
MARL extends classical reinforcement learning and Markov decision processes to settings where multiple agents learn and act simultaneously~\cite{sutton2018reinforcement, puterman2014markov, tsitsiklis1997analysis, tan1993multi}. 
In cooperative tasks, CTDE is widely adopted to mitigate partial observability and non-stationarity by using centralized information during training while preserving decentralized execution~\cite{lowe2017multi, sunehag2017value}. 
Value-decomposition methods, including VDN, QMIX, Weighted QMIX, QTRAN, and QPLEX, factorize the joint value into agent-wise utilities under different structural constraints~\cite{sunehag2017value, rashid2018qmix, rashid2020weighted, son2019qtran, wang2020qplex, hu2021riit}. 
Policy-gradient and actor-critic methods such as COMA, MADDPG, FACMAC, IPPO, MAPPO, HAPPO, and MAT further improve optimization stability and scalability in cooperative games~\cite{foerster2018counterfactual, lowe2017multi, peng2021facmac, de2020independent, de2020IPPO, yu2022surprising, kuba2022trust, wen2022MAT, wen2022multi}. 
HySTAR builds on MAPPO and augments the critic with structured high-order value decomposition in place of a single unstructured global value.

\subsection{Graph and Hypergraph Representation Learning}
Graph neural networks have been widely used in MARL to model relational interactions through graph convolution, relevance modeling, and pairwise message passing~\cite{kipf2017semi,jiang2018graph,malysheva2018deep}.
Pairwise graphs represent binary relations, whereas coordinated behaviors often involve coalitions of more than two agents.
Hypergraph learning directly represents such non-binary relations through hyperedges~\cite{hammond2011wavelets,feng2019hypergraph,yadati2019hypergcn,bai2021hypergraph}.

\begin{figure*}[t]
    \centering
    \includegraphics[width=1\linewidth]{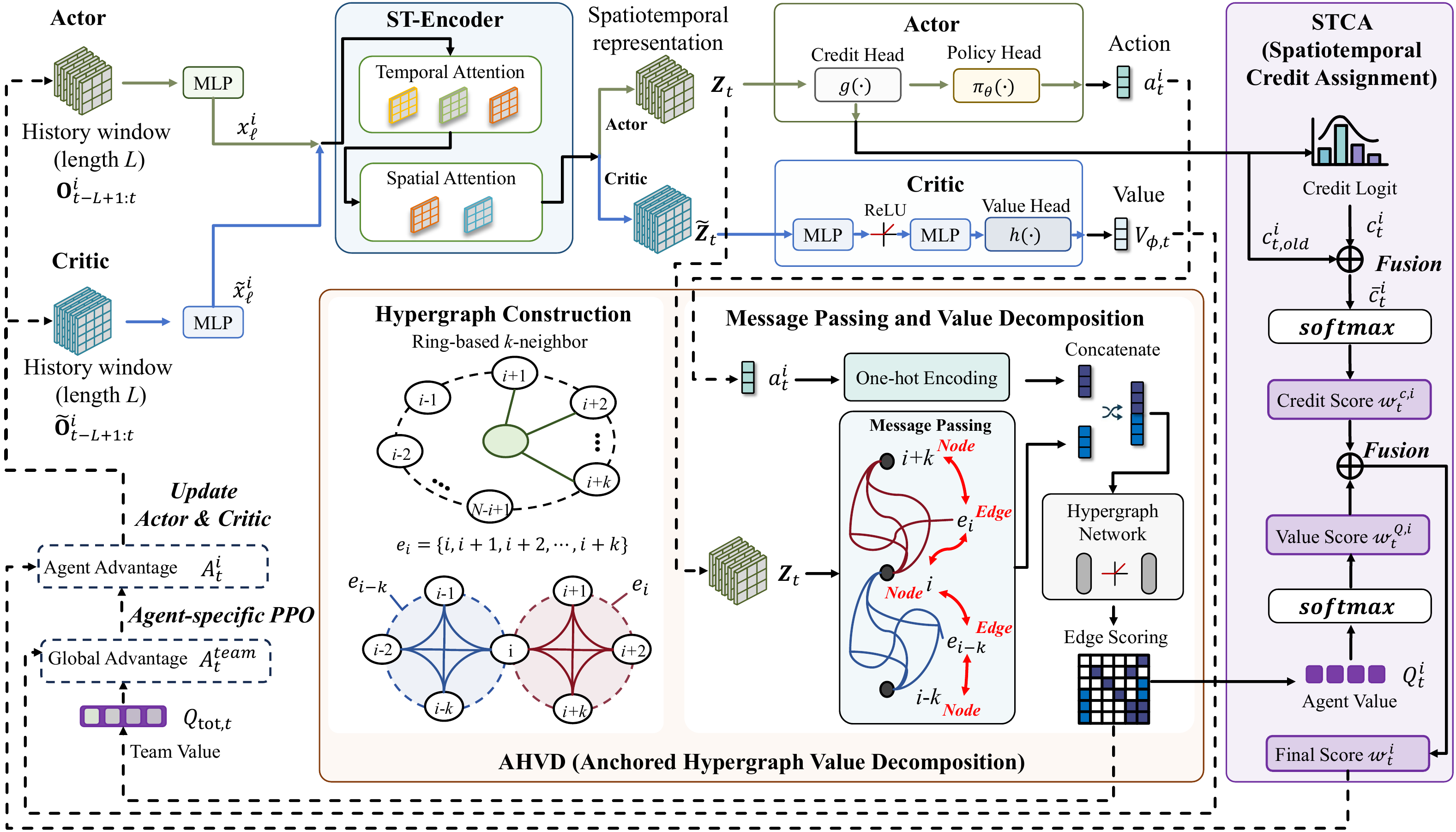}
    \caption{
Overall architecture of HySTAR.
The actor and centralized critic use independent MLP/ST-Encoder branches
with the same architecture: their MLP outputs are $x_t^i$ and
$\widetilde{x}_t^i$, and their spatiotemporal representations are $Z_t^i$
and $\widetilde{Z}_t^i$, respectively.
The Policy Head $\pi_{\theta}(\cdot)$ outputs $a_t^i$, the Credit Head
$g(\cdot)$ outputs the Credit Logit $c_t^i$, and the Critic Value Head
$h(\cdot)$ outputs $V_{\phi,t}$.
During training, AHVD propagates information through overlapping routes
$i-k\rightarrow e_{i-k}\rightarrow i\rightarrow e_i\rightarrow i+k$.
STCA fuses the resulting Credit and Value Scores into Final Scores for
agent-specific policy advantages.}
    \label{fig:pipeline}
\end{figure*}

HYGMA combines dynamic spectral clustering with hypergraph convolution to discover adaptive agent groups and generate group-aware representations for both value-based and policy-based MARL~\cite{liu2025hygma}.
Its policy-based implementation allows the actor to use structured information aggregated from agents in the same group.
HySTAR adopts a comparable communication-augmented execution setting and investigates a different structural hypothesis:
HySTAR maintains an anchored sparse hypergraph skeleton and adapts the representations propagated over it.
This contrast highlights the difference between dynamic structural adaptation and a stable decomposition basis under comparable communication-augmented execution settings.

\subsection{Value Decomposition and Credit Assignment}
Credit assignment remains a central challenge in shared-reward MARL. 
Counterfactual and Shapley-style methods derive individual contribution signals from alternative actions or marginal utilities~\cite{foerster2018counterfactual, wang2020sqddpg}, while implicit credit assignment and exploration-driven approaches improve learning signals through auxiliary objectives or behavioral diversity~\cite{zhou2020learning, yang2020multi, zheng2022episodic}. 
Reward shaping provides another route to guide policy optimization while preserving desirable optimality properties~\cite{ng1999policy, devlin2012dynamic}. 
STCA complements these lines of work by combining temporal credit information with high-order structural value scores and converting the shared team advantage into agent-specific shaped advantages.

\section{Method}
\label{sec:method}

We model a cooperative task as a Decentralized Partially Observable
Markov Decision Process (Dec-POMDP)
$\mathcal{G}=\langle\mathcal{N},\mathcal{S},\mathcal{A},
P,R,\Omega,O,\gamma\rangle$.
Here, $\mathcal{N}=\{1,\ldots,N\}$ is the agent set,
$s_t\in\mathcal{S}$ is the environment state, agent $i$ receives
$o_t^i\in\Omega$, and the joint action $\mathbf{a}_t$ yields
$s_{t+1}\sim P(\cdot\mid s_t,\mathbf{a}_t)$ and shared reward
$r_t=R(s_t,\mathbf{a}_t)$.

Let $\mathbf{O}_{t-L+1:t}^i=(o_{t-L+1}^i,\ldots,o_t^i)$ denote the
length-$L$ history of agent $i$, and let
$\mathbf{O}_{t-L+1:t}=\{\mathbf{O}_{t-L+1:t}^i\}_{i=1}^{N}$ collect all
agent histories. The centralized critic analogously uses
$\widetilde{\mathbf{O}}_{t-L+1:t}
=\{\widetilde{o}_\ell^i\}_{\ell=t-L+1:t,\,i=1}^{N}$.
HySTAR follows communication-augmented CTDE because the actor ST-Encoder
aggregates all agent histories during action generation. Its actor branch
produces
\begin{equation}
\begin{aligned}
\mathbf{Z}_t
&=f_{\theta}^{\mathrm{ST}}\!\left(\mathbf{O}_{t-L+1:t}\right),\\
a_t^i&\sim\pi_{\theta}(\cdot\mid Z_t^i),
\qquad c_t^i=g(Z_t^i).
\end{aligned}
\end{equation}
Here $\mathbf{Z}_t=[Z_t^1,\ldots,Z_t^N]^\top$. The independent critic
branch produces
\begin{equation}
\widetilde{\mathbf{Z}}_t
=f_{\phi}^{\mathrm{ST}}\!\left(\widetilde{\mathbf{O}}_{t-L+1:t}\right),
\qquad
V_{\phi,t}=h(\widetilde{\mathbf{Z}}_t).
\end{equation}

\subsection{Overview}
HySTAR follows the principle of
\emph{anchor the topology, adapt the representations}.
As shown in Fig.~\ref{fig:pipeline}, the actor branch outputs actions and
Credit Logits, while the baseline critic provides $V_{\phi,t}$ for GAE.
AHVD evaluates action-conditioned coordination values on a fixed,
overlapping hypergraph, and STCA combines its Value Scores with the
Credit Scores to allocate the mixed team advantage across agents.
AHVD, STCA, target networks, and centralized observations are used only
during training. 

\subsection{Spatiotemporal Representation Learning}
\label{subsec:st_encoder}

At each timestep $\ell$, the actor and critic MLPs in Fig.~\ref{fig:pipeline}
encode their inputs as
\begin{equation}
 x_\ell^i=f_{\mathrm{enc}}^{\pi}(o_\ell^i),
 \qquad
 \widetilde{x}_\ell^i=f_{\mathrm{enc}}^{V}(\widetilde{o}_\ell^i),
\end{equation}
so $x_t^i$ and $\widetilde{x}_t^i$ are the final-position MLP outputs
shown in the figure. Each ST-Encoder applies causal temporal
self-attention along each agent history, followed by spatial
self-attention across agents at the same timestep. Residual fusion gives
\begin{equation}
 Z_\ell^i
 =\mathrm{LN}\!\left(
 f_{\mathrm{fuse}}(x_{\ell,\mathrm{tem}}^i+x_{\ell,\mathrm{spa}}^i)
 \right),
\end{equation}
and the critic uses the same architecture with independent parameters to
produce $\widetilde{Z}_\ell^i$ from $\widetilde{x}_\ell^i$.
The final-position stacks $\mathbf{Z}_t$ and
$\widetilde{\mathbf{Z}}_t$ feed the actor and critic heads, respectively;
$\mathbf{Z}_t$ is also stored with the rollout for AHVD optimization.

\subsection{Anchored Hypergraph Value Decomposition}
\label{subsec:ahvd}

AHVD keeps the sparse decomposition support fixed while learning the
propagated messages and action-aware value scores. For each agent slot
$i$, its ring neighborhood and anchored hyperedge are
\begin{equation}
\mathcal{N}(i)=\{i+1,\ldots,i+k\},
\qquad
e_i=\{i\}\cup\mathcal{N}(i),
\end{equation}
where indices wrap around $\{1,\ldots,N\}$. The support is constructed
once before training, giving uniformly covered overlapping hyperedges
with $Nk$ directed incidences. 

AHVD initializes $h_i^{(0)}=W_pZ_t^i+b_p$ and applies the compact
node--edge--node update
\begin{align}
m_{e_i}^{(\ell)}
&=\frac{1}{|\mathcal{N}(i)|}
  \sum_{j\in\mathcal{N}(i)}h_j^{(\ell)},\\
h_i^{(\ell+1)}
&=\mathrm{LN}\!\left(
 h_i^{(\ell)}+
 \mathrm{MLP}_{\ell}[h_i^{(\ell)}\parallel m_{e_i}^{(\ell)}]
 \right).
\end{align}
Because adjacent hyperedges overlap, stacked layers realize routes such
as $i-k\rightarrow e_{i-k}\rightarrow i\rightarrow e_i\rightarrow i+k$
with $\mathcal{O}(Nk)$ message complexity. After the final layer, let
$h_i$ denote the propagated feature. For each $j\in\mathcal{N}(i)$, the
edge-scoring network computes
\begin{equation}
q_{ij,t}=f_{\psi}\!\left[
 h_i\parallel h_j\parallel
 \mathrm{onehot}(a_t^i)\parallel\mathrm{onehot}(a_t^j)
\right].
\end{equation}
The agent and team values are
\begin{equation}
Q_t^i=\sum_{j\in\mathcal{N}(i)}q_{ij,t},
\qquad
Q_{\mathrm{tot},t}=\frac{1}{2}\sum_{i=1}^{N}Q_t^i,
\end{equation}
where $1/2$ follows the symmetric-edge implementation.

The trainer normalizes $Q_{\mathrm{tot},t}$ within the minibatch and
$\{Q_t^i\}_{i=1}^{N}$ across agents; denote the results by
$\bar Q_{\mathrm{tot},t}$ and $\bar Q_t^i$. The AHVD team advantage is
\begin{equation}
\label{eq:team_adv}
A_t^{\mathrm{team}}
=\mathrm{Norm}\!\left(
\bar Q_{\mathrm{tot},t}-\operatorname{sg}[V_{\phi,t}]
\right),
\end{equation}
where $\operatorname{sg}[\cdot]$ denotes stop-gradient. With normalized
GAE advantage $A_t^{\mathrm{base}}$ and update-dependent ramp
$\rho_u\in[0,1]$, HySTAR forms
\begin{equation}
\label{eq:mixed_adv}
A_t^{\mathrm{mix}}
=(1-\rho_u\eta_{\mathrm{adv}})A_t^{\mathrm{base}}
+\rho_u\eta_{\mathrm{adv}}A_t^{\mathrm{team}},
\end{equation}
then normalizes it before STCA.

\subsection{Spatiotemporal Credit Assignment}
\label{subsec:stca}

For every PPO minibatch, the Credit Head $g(\cdot)$ recomputes the
current Credit Logit $c_t^i$ and retrieves the rollout Credit Logit
$c_{t,\mathrm{old}}^i$ from the buffer.
The implementation uses the equal-weight logit mixture
\begin{equation}
\bar c_t^i
=
\frac{1}{2}c_{t,\mathrm{old}}^i
+
\frac{1}{2}c_t^i.
\end{equation}
A temperature-scaled softmax converts the mixed Credit Logit into the
Credit Score
\begin{equation}
w_t^{c,i}
=
\frac{\exp(\bar c_t^i/T_c)}
{\sum_{j=1}^{N}\exp(\bar c_t^j/T_c)},
\end{equation}
where $T_c$ is the credit temperature.

\begin{figure*}[t]
\centering
\includegraphics[width=\textwidth]{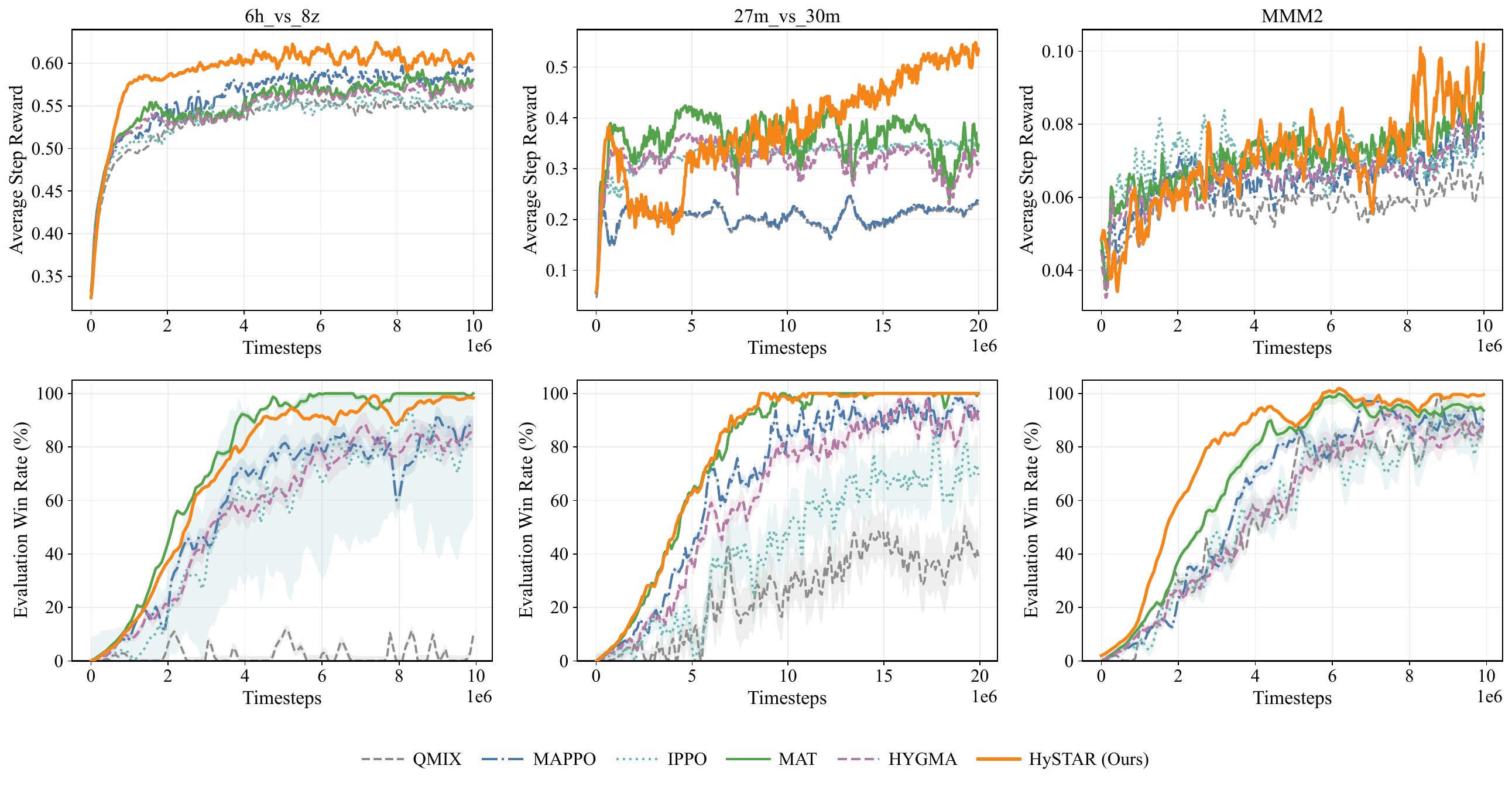}
\caption{
Average step reward (top) and evaluation win rate (bottom) on three
super-hard SMAC maps: \texttt{6h\_vs\_8z},
\texttt{27m\_vs\_30m}, and \texttt{MMM2}.
Curves and shaded regions denote the mean and one standard deviation
across independent seeds.
}
\label{fig:reward_curves_hard_maps}
\end{figure*}

\begin{table*}[t]
\centering
\caption{
Median evaluation win rate and standard deviation on selected SMAC maps.
Difficulty is indicated for each map.
For each map, we bold the method with the highest median win rate;
when multiple methods have the same median, the one with the lower
standard deviation is highlighted.
Methods with identical median and standard deviation are jointly highlighted.
}
\label{tab:win_rate_final}

\resizebox{\textwidth}{!}{
\begin{tabular}{ll|cccccc}
\toprule
\textbf{Map}
& \textbf{Difficulty}
& \textbf{QMIX}~\cite{rashid2018qmix}
& \textbf{IPPO}~\cite{de2020IPPO}
& \textbf{MAPPO}~\cite{yu2022surprising}
& \textbf{MAT}~\cite{wen2022MAT}
& \textbf{HYGMA}~\cite{liu2025hygma}
& \textbf{HySTAR (Ours)} \\
\midrule

3m
& Easy
& 96.9\err{1.3}
& \best{100.0}{0.0}
& \best{100.0}{0.0}
& 100.0\err{1.8}
& 99.1\err{1.1}
& \best{100.0}{0.0} \\

1c3s5z
& Easy
& 96.1\err{1.7}
& \best{100.0}{0.0}
& \best{100.0}{0.0}
& 100.0\err{2.4}
& 97.0\err{2.0}
& \best{100.0}{0.0} \\

8m
& Easy
& 97.7\err{1.9}
& 100.0\err{0.7}
& \best{100.0}{0.0}
& 100.0\err{1.1}
& 98.0\err{1.6}
& \best{100.0}{0.0} \\

MMM
& Easy
& 95.3\err{2.5}
& 96.9\err{0.0}
& 96.9\err{0.6}
& 100.0\err{2.2}
& 98.2\err{2.9}
& \best{100.0}{1.8} \\

3s\_vs\_5z
& Hard
& 98.4\err{2.4}
& \best{100.0}{0.0}
& 100.0\err{0.6}
& 100.0\err{1.7}
& 99.0\err{0.9}
& \best{100.0}{0.0} \\

5m\_vs\_6m
& Hard
& 75.8\err{3.7}
& 87.5\err{2.3}
& 89.1\err{2.5}
& 93.8\err{4.4}
& 95.2\err{2.3}
& \best{95.6}{2.1} \\

8m\_vs\_9m
& Hard
& 92.2\err{2.0}
& 96.9\err{0.7}
& 96.9\err{0.6}
& 100.0\err{3.1}
& 92.0\err{3.2}
& \best{100.0}{0.9} \\

3s5z
& Hard
& 88.3\err{2.9}
& 96.9\err{1.5}
& 96.9\err{0.7}
& 100.0\err{1.9}
& 88.0\err{3.8}
& \best{100.0}{0.9} \\

3s5z\_vs\_3s6z
& Super Hard
& 82.8\err{5.3}
& 82.8\err{19.1}
& 84.4\err{34.0}
& 97.2\err{1.3}
& 88.6\err{3.4}
& \best{98.5}{0.9} \\

6h\_vs\_8z
& Super Hard
& 9.4\err{2.0}
& 84.4\err{33.3}
& 88.3\err{3.7}
& \best{100.0}{1.3}
& 85.0\err{5.2}
& 98.3\err{0.9} \\

27m\_vs\_30m
& Super Hard
& 39.1\err{9.8}
& 69.5\err{11.8}
& 93.8\err{2.4}
& 100.0\err{0.7}
& 90.0\err{4.1}
& \best{100.0}{0.4} \\

MMM2
& Super Hard
& 87.5\err{2.6}
& 86.7\err{7.3}
& 90.6\err{2.8}
& 93.6\err{2.6}
& 88.0\err{4.6}
& \best{99.6}{1.2} \\

\bottomrule
\end{tabular}
}
\end{table*}

\begin{figure*}[t]
    \centering
    \includegraphics[width=\textwidth]{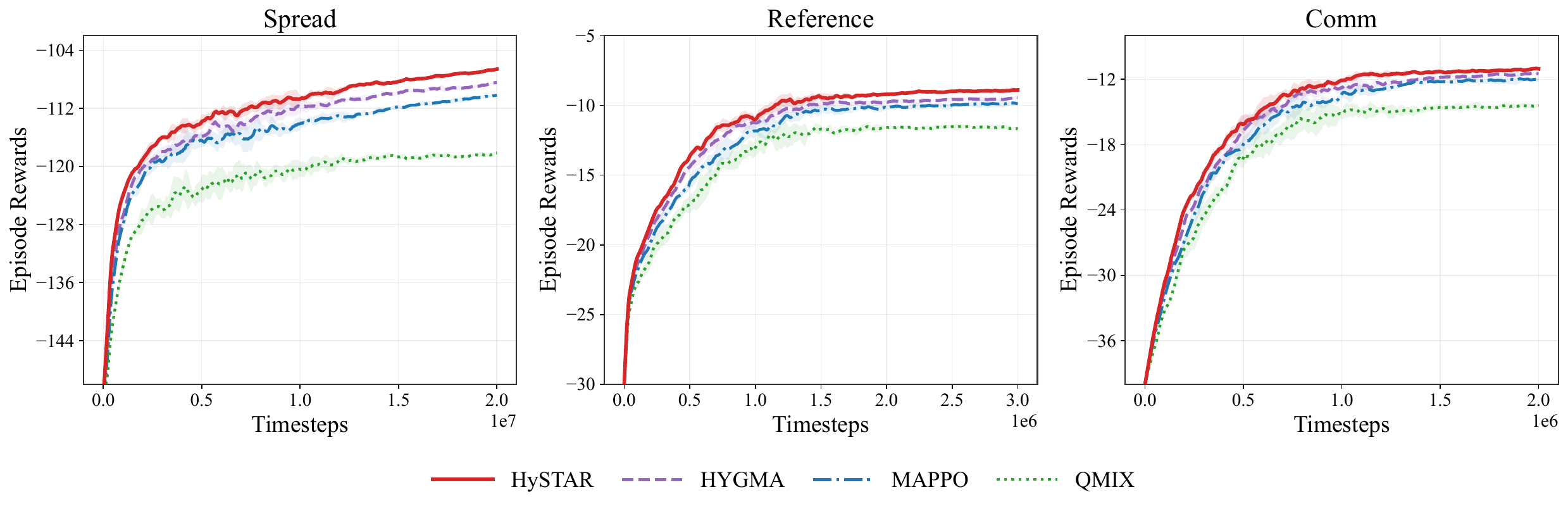}
    \caption{
    Learning curves on the cooperative MPE tasks Spread, Reference, and Comm.
    Curves and shaded regions denote the mean and one standard deviation across
    random seeds; higher episode reward is better.
    }
    \label{fig:mpe_reward_curves}
\end{figure*}
\begin{table*}[t]
\centering
\caption{
Average success rate and standard deviation on six GRF academy scenarios.
}
\label{tab:grf_success_rate}
\resizebox{\textwidth}{!}{
\begin{tabular}{l|cccccc}
\toprule
\textbf{Scenario}
& \textbf{QMix}~\cite{rashid2018qmix}
& \textbf{CDS}~\cite{chenghao2021CDS}
& \textbf{TiKick}~\cite{huang2021tikick}
& \textbf{MAPPO}~\cite{yu2022surprising}
& \textbf{HYGMA}~\cite{liu2025hygma}
& \textbf{HySTAR (Ours)} \\
\midrule

3v.1
& 8.12\err{2.83}
& 76.60\err{3.27}
& 76.88\err{3.15}
& 88.03\err{1.06}
& {97.70}\err{0.70}
& \best{98.26}{0.93}\\

CA(easy)
& 15.98\err{2.85}
& 63.28\err{4.89}
& /
& 87.76\err{1.34}
& 90.12\err{1.26}
& \best{91.80}{1.05} \\

CA(hard)
& 3.22\err{1.60}
& 58.35\err{5.56}
& 73.09\err{2.08}
& 77.38\err{4.81}
& 69.67\err{5.21}
& \best{84.60}{3.20}\\

Corner
& 16.10\err{3.00}
& 3.80\err{0.54}
& 33.00\err{3.01}
& 65.53\err{2.19}
& 56.18\err{3.66}
& \best{74.90}{2.50}\\

PS
& 8.05\err{3.66}
& 94.15\err{2.54}
& /
& 94.92\err{0.68}
& 90.36\err{0.73}
& \best{96.30}{0.70}\\

RPS
& 8.08\err{4.71}
& 62.38\err{4.56}
& 79.12\err{2.06}
& 76.83\err{1.81}
& 63.11\err{4.04}
& \best{84.20}{2.40} \\

\bottomrule
\end{tabular}
}
\end{table*}

\begin{table}[t]
\centering
\caption{
Traffic Junction success rate and convergence epochs.
}
\label{tab:traffic_junction}

\setlength{\tabcolsep}{3.5pt}
\renewcommand{\arraystretch}{1.10}

\resizebox{\columnwidth}{!}{
\begin{tabular}{lcc}
\toprule
\textbf{Method}
&
\begin{tabular}[c]{@{}c@{}}
$\mathbf{7\times7}$, $N_{\max}=5$\\
$p_{\mathrm{arrive}}=0.3$
\end{tabular}
&
\begin{tabular}[c]{@{}c@{}}
$\mathbf{14\times14}$, $N_{\max}=10$\\
$p_{\mathrm{arrive}}=0.2$
\end{tabular}
\\
\midrule

TarMAC-IC3Net~\cite{das2019tarmac}
& $84.8 \pm 4.5\%~(599 \pm 187)$
& $95.5 \pm 1.3\%~(1706 \pm 104)$
\\

GA-Comm~\cite{liu2020gacomm}
& $95.9 \pm 0.1\%~(891 \pm 141)$
& $97.1 \pm 0.7\%~(1573 \pm 253)$
\\

MAGIC~\cite{niu2021magic}
& $99.9 \pm 0.1\%~(440 \pm 64)$
& $99.9 \pm 0.1\%~(819 \pm 85)$
\\

HYGMA~\cite{liu2025hygma}
& $99.7 \pm 0.1\%~(272 \pm 41)$
& $99.2 \pm 0.1\%~(569 \pm 34)$
\\

\midrule

\textbf{HySTAR (Ours)}
& \textbf{$99.9 \pm 0.1\%~(263 \pm 45)$}
& \textbf{$99.9 \pm 0.1\%~(558 \pm 56)$}
\\

\bottomrule
\end{tabular}
}
\end{table}

A temperature-scaled softmax converts the normalized AHVD agent values
into the Value Score
\begin{equation}
w_t^{Q,i}
=
\frac{\exp(\bar Q_t^i/T_c)}
{\sum_{j=1}^{N}\exp(\bar Q_t^j/T_c)}.
\end{equation}
The update-dependent structural mixing coefficient is
$\rho_u\eta_{\mathrm{credit}}$.
STCA fuses the Credit Score and Value Score into the Final Score
\begin{equation}
w_t^i
=
(1-\rho_u\eta_{\mathrm{credit}})w_t^{c,i}
+
\rho_u\eta_{\mathrm{credit}}w_t^{Q,i},
\qquad
\sum_{i=1}^{N}w_t^i=1.
\end{equation}
The agent-specific PPO advantage is
\begin{equation}
\label{eq:stca_adv}
A_t^i
=
N\,
\operatorname{sg}[w_t^i]\,
A_t^{\mathrm{mix}}.
\end{equation}
Since $\sum_i w_t^i=1$, the factor $N$ makes the mean agent advantage
equal to $A_t^{\mathrm{mix}}$, preserving the shared-advantage scale.
The stop-gradient operation is applied to the Final Score after score
fusion, so the PPO loss treats $w_t^i$ as a fixed minibatch allocation
coefficient.
The actor representation and policy head are optimized by the PPO
objective, and the credit logits are recomputed from the updated actor
representation in subsequent minibatches.
Further implementation details are provided in the supplementary
material.

\subsection{Optimization Objective}
\label{subsec:objective}

HySTAR uses three separately optimized objectives.
The actor follows the PPO clipped objective with the STCA advantage
from Eq.~\eqref{eq:stca_adv}:
\begin{equation}
\begin{aligned}
\mathcal{J}_{\pi}(\theta)
=
\mathbb{E}_t\Big[
\min\big(
& r_t^i(\theta)A_t^i,\\
& \mathrm{clip}
\left(
r_t^i(\theta),1-\epsilon,1+\epsilon
\right)A_t^i
\big)
+
\beta\mathcal{H}(\pi_\theta)
\Big].
\end{aligned}
\end{equation}
Here, $r_t^i(\theta)$ is the PPO policy ratio, $\epsilon$ is the
clipping parameter, and $\beta$ is the entropy coefficient.

The baseline critic is trained against the return target $\widehat R_t$
with the standard clipped MAPPO value loss:
\begin{equation}
\mathcal{L}_{V}(\phi)
=
\mathbb{E}_t
\left[
\ell_{\mathrm{clip}}
\left(
V_{\phi,t},
V_{\phi,t}^{\mathrm{old}},
\widehat R_t
\right)
\right].
\end{equation}

AHVD uses a target actor and a target AHVD critic.
For the mean team reward $r_t^{\mathrm{team}}$, the TD target is
\begin{equation}
y_t^{Q}
=
r_t^{\mathrm{team}}
+
\gamma m_{t+1}^{\mathrm{team}}
Q_{\mathrm{tot}}^{\mathrm{tar}}
\left(
\mathbf{Z}_{t+1}^{\mathrm{tar}},
\mathbf{a}_{t+1}^{\mathrm{tar}}
\right),
\end{equation}
and the AHVD loss is
\begin{equation}
\mathcal{L}_{\mathrm{AHVD}}(\psi)
=
\mathbb{E}_t
\left[
\left(
Q_{\mathrm{tot},t}-y_t^{Q}
\right)^2
\right].
\end{equation}
The actor, baseline critic, and AHVD critic use separate optimizers.
The target actor and target AHVD critic follow the online networks
through soft updates.
Crucially, to prevent representation collapse caused by the stop-gradient operation in Eq. (16), the Credit Head $g(\cdot)$ is actively trained via an auxiliary distillation loss $\mathcal{L}_{credit}$. This objective explicitly aligns the temporal credit logits $c_t^i$ with the structural agent values $Q_t^i$ evaluated by AHVD. This decoupled optimization stabilizes the primary policy update while guaranteeing end-to-end learning of the spatiotemporal representations (detailed in Appendix D.1).

\section{Experiments}
\label{sec:experiments}

\subsection{Experimental Setup}
We evaluate HySTAR on SMAC~\cite{vinyals2019SCMA}, the multi-agent particle-world environment (MPE)~\cite{lowe2017multi}, Google Research Football (GRF)~\cite{kurach2020google}, and Traffic Junction~\cite{sukhbaatar2016commnet}. SMAC stresses partially observable combat with heterogeneous units and agent death. MPE evaluates particle-world coordination on Spread, Reference, and Comm. Following the MAPPO setup, local observations are concatenated as the centralized state, and Comm uses separate parameters for its heterogeneous agents. GRF evaluates coordinated play against scripted opponents. Traffic Junction contains stochastic vehicle arrivals and a varying active-agent set, and we follow the HYGMA settings $7\times7$ with $N_{\max}=5$, $p_{\mathrm{arrive}}=0.3$, and $14\times14$ with $N_{\max}=10$, $p_{\mathrm{arrive}}=0.2$~\cite{liu2025hygma}. We report median win rate on SMAC, success rate on GRF, success rate with convergence epochs on Traffic Junction, and episode reward on MPE; learning curves show the mean and one standard deviation across seeds. Unless otherwise noted, $\lambda_c=0.5$, $\eta=0.5$, $L=10$. Full hyperparameters, per-environment configurations, and the number of seeds used for each benchmark are given in Appendix~A.
The main SMAC comparison covers monotonic value factorization (QMIX~\cite{rashid2018qmix}), independent policy optimization (IPPO~\cite{de2020IPPO}), centralized policy optimization (MAPPO~\cite{yu2022surprising}), transformer-based coordination (MAT~\cite{wen2022MAT}), and dynamic hypergraph grouping (HYGMA~\cite{liu2025hygma}).
We select methods whose reported results or implementations align with the selected map set and evaluation format; Appendix~A documents the source and protocol of each baseline result.

\subsection{Performance on SMAC}
Tab.~\ref{tab:win_rate_final} and Fig.~\ref{fig:reward_curves_hard_maps} summarize the SMAC results. HySTAR achieves the best reported result on 11 of 12 maps. On \texttt{3s5z\_vs\_3s6z}, it improves over MAPPO from $84.4\%$ to $98.5\%$, corresponding to $+14.1$ percentage points and a 16.7\% relative gain. On \texttt{6h\_vs\_8z}, it improves over HYGMA from $85.0\%$ to $98.3\%$ ($+13.3$ points; 15.6\% relative). The reward and win-rate curves further show faster and more stable learning on the three super-hard maps. The topology and agent-death analyses separately examine masking, topology dynamics, sparse high-order aggregation, and critic stability.

\subsection{Performance on MPE}
Fig.~\ref{fig:mpe_reward_curves} compares HySTAR with HYGMA, MAPPO and QMIX on Spread, Reference, and Comm. HySTAR obtains the highest final reward on all three tasks, with the clearest advantage on Spread, where agents must form complementary spatial roles for landmark coverage. Because MPE returns are negative, percentage ratios reverse the usual sign convention; we therefore report absolute reward improvements. The fixed active-agent set also shows that anchoring is useful beyond agent-death or variable-team settings: it supplies a stable value-decomposition basis while the ST-Encoder adapts task-dependent interactions.

\subsection{Performance on GRF}
Tab.~\ref{tab:grf_success_rate} reports the results under a common interaction and evaluation budget. HySTAR ranks first on all six scenarios. On \texttt{Corner}, it raises success rate from $65.53\%$ with MAPPO to $74.90\%$ ($+9.37$ points; 14.3\% relative), while on \texttt{RPS} it improves over HYGMA from $63.11\%$ to $84.20\%$ ($+21.09$ points; 33.4\% relative). The larger gains on demanding role-coordination tasks support the transfer of anchored high-order credit assignment beyond SMAC.

\subsection{Performance on Traffic Junction}
As shown in Tab.~\ref{tab:traffic_junction}, HySTAR reaches $99.9\%$ success in both settings. Relative to HYGMA, it reduces convergence epochs by 3.3\% ($272\!\rightarrow\!263$) and 1.9\% ($569\!\rightarrow\!558$); relative to MAGIC, the reductions are 40.2\% and 31.9\%. Since stochastic arrivals continuously change the active-agent set, these gains support the intended mechanism: the anchored topology preserves the decomposition basis of the critic while node representations adapt to current traffic.

\begin{figure}[t]
    \centering
    \includegraphics[width=\linewidth]{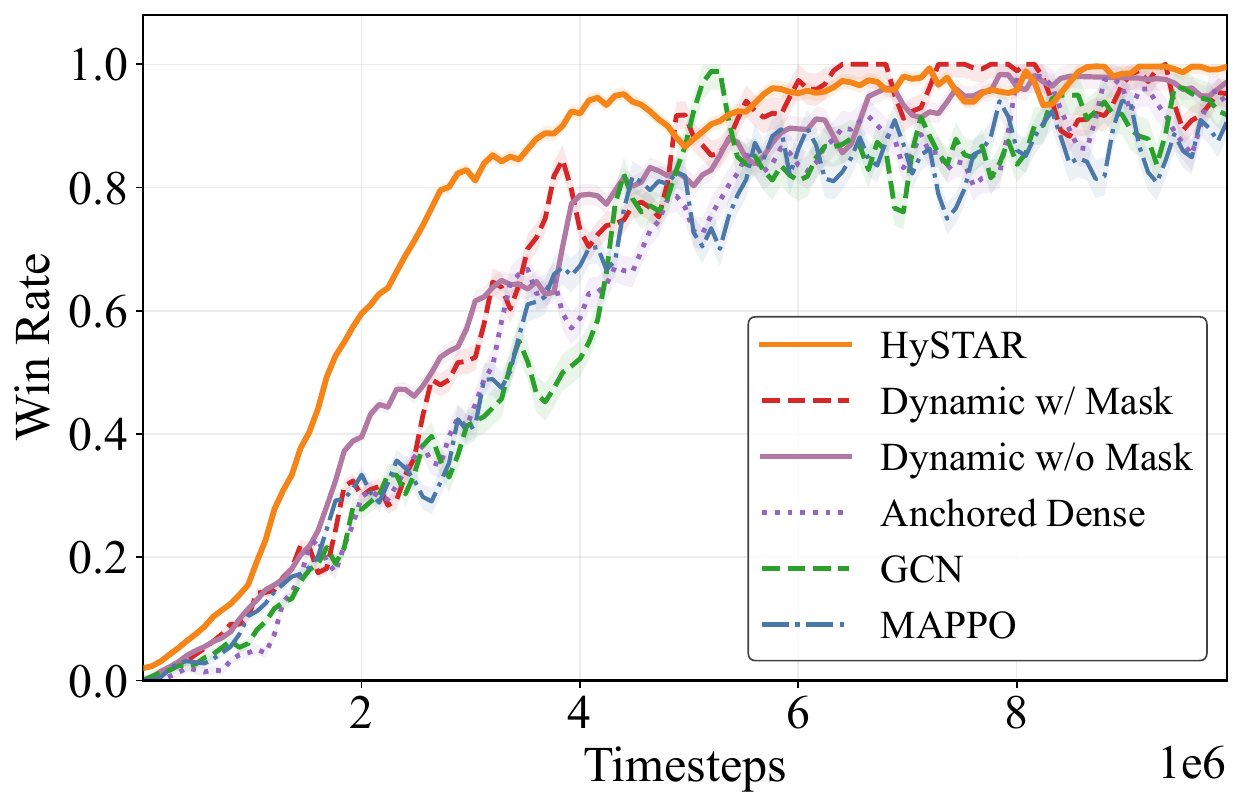}
    \caption{
    Topology ablation on \texttt{MMM2}.
    All hypergraph and graph variants use the same ST-Encoder, actor input,
    representation dimension, training budget, and evaluation protocol;
    MAPPO is included as an external reference.
    HySTAR achieves the highest and most stable win rate.
    }
    \label{fig:topology_ablation}
\end{figure}

\begin{figure}[t]
    \centering
    \includegraphics[width=\linewidth]{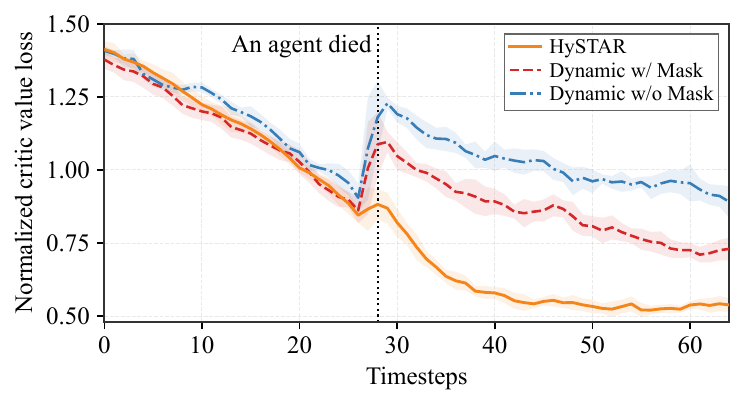}
    \caption{ Normalized critic value loss around an agent-death event on \texttt{MMM2}.
    }
    \label{fig:agent_death_value_loss}
\end{figure}

\begin{figure}[t]
    \centering
    \includegraphics[width=\linewidth]{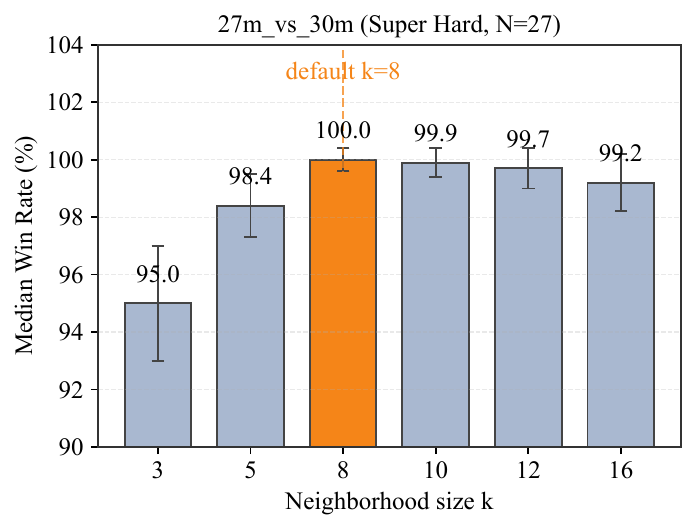}
    \caption{ Sensitivity to the neighborhood size $k$ on the large-scale
\texttt{27m\_vs\_30m} scenario.
    }
    \label{fig:k_sensitivity}
\end{figure}

\begin{table}[t]
\centering
\caption{Parameter efficiency on \texttt{27m\_vs\_30m}.}
\label{tab:param_efficiency_27m}
\resizebox{\columnwidth}{!}{
\begin{tabular}{l|cc|cc}
\toprule
\textbf{Method}
& \textbf{Params (K)}
& \textbf{$\Delta$ Param.}
& \textbf{Win Rate}
& \textbf{$\Delta$ Win}
\\
\midrule
MAPPO~\cite{yu2022surprising}
& 125.861 & 0.0\% & 93.8\% & 0.0\% \\
IPPO~\cite{de2020IPPO}
& \textbf{114.641} & \textbf{+8.9\%}
& 69.5\% & $-25.9\%$ \\
MAT~\cite{wen2022MAT}
& 253.455 & $-101.4\%$
& \textbf{100.0\%} & \textbf{+6.6\%} \\
HySTAR (Ours)
& 149.889 & $-19.1\%$
& \textbf{100.0\%} & \textbf{+6.6\%} \\
\bottomrule
\end{tabular}
}
\end{table}

\subsection{Ablation Study}
\label{subsec:ablation}

\paragraph{Topology.}
Fig.~\ref{fig:topology_ablation} controls the representation backbone and actor input across the hypergraph and graph variants. The comparison evaluates critic topology and aggregation under the same spatiotemporal features. Masking improves dynamic reconstruction, while both dynamic variants remain below HySTAR. Anchored Dense and GCN also underperform; these results separate the benefit of the anchored sparse scaffold from additional connectivity and pairwise aggregation. MAPPO serves as an external policy baseline, and the structural variants provide the representation-controlled comparison.

\paragraph{Agent death.}
Fig.~\ref{fig:agent_death_value_loss} aligns normalized critic loss around an allied-agent death. Dynamic w/o Mask shows the largest increase and slowest recovery, while Dynamic w/ Mask is intermediate. HySTAR exhibits the smallest disturbance and lowest late-stage loss, showing that anchoring stabilizes the critic beyond masking alone.

\paragraph{Neighborhood size.}
Fig.~\ref{fig:k_sensitivity} varies $k$ on \texttt{27m\_vs\_30m}. Small neighborhoods omit useful coordination context, whereas values above $k=8$ add redundant interactions and yield unchanged or lower performance.

\subsection{Parameter Efficiency Analysis}
\label{subsec:param_efficiency}
Tab.~\ref{tab:param_efficiency_27m} compares critic capacity on \texttt{27m\_vs\_30m}. HySTAR reaches a $100.0\%$ win rate with 149.889K critic parameters, using 40.9\% fewer parameters than MAT while matching its win rate. Compared with MAPPO, HySTAR uses 19.1\% more critic parameters but improves win rate from $93.8\%$ to $100.0\%$ (6.6\% relative), yielding a favorable performance--capacity trade-off.

\subsection{Limitations}
\label{subsec:limitations}
HySTAR assumes a consistent agent-slot ordering, and general permutations produce a different anchored incidence pattern. Execution uses cross-agent representations and therefore follows communication-augmented CTDE. The ST-Encoder, AHVD, and STCA add critic parameters relative to MAPPO and IPPO. Additional scope and implementation details are provided in the supplementary material.

\section{Conclusion}
\label{sec:conclusion}

We presented HySTAR, a MAPPO-based framework
that combines an anchored sparse decomposition scaffold with adaptive
spatiotemporal agent representations.
AHVD performs high-order value decomposition over this temporally
consistent support, while STCA produces agent-specific optimization
signals.
Across SMAC, GRF, Traffic Junction, and MPE, HySTAR consistently
outperforms MAPPO-style, value-factorization, and dynamic-grouping
baselines.
Controlled topology variants and the agent-death analysis support
anchoring a sparse decomposition scaffold while adapting the
representations propagated over it.

\bibliography{aaai2027}


\end{document}